\documentclass{article} 
\usepackage{iclr2027_conference,times}

\usepackage{amsmath,amsfonts,bm}

\def\eqref#1{equation~\ref{#1}}

\def\1{\bm{1}}

\DeclareMathAlphabet{\mathsfit}{\encodingdefault}{\sfdefault}{m}{sl}
\SetMathAlphabet{\mathsfit}{bold}{\encodingdefault}{\sfdefault}{bx}{n}

\usepackage{textcomp}
\usepackage{stfloats}
\usepackage{url}
\usepackage{verbatim}
\usepackage{graphicx}
\usepackage{multirow}
\usepackage{xspace}
\usepackage{bbm}
\usepackage{booktabs}
\usepackage{color}
\usepackage{xcolor}
\usepackage{setspace}
\usepackage{braket}
\usepackage{tablefootnote}
\usepackage{subcaption}
\usepackage[normalem]{ulem}
\usepackage{hyperref}
\usepackage{cleveref}
\usepackage{makecell}
\usepackage{wrapfig}
\usepackage{sansmath}
\usepackage{float}
\usepackage[table]{xcolor}
\usepackage[most]{tcolorbox}
\usepackage{listings}
\usepackage{tabularx}
\usepackage{xspace}
\usepackage{wrapfig}
\usepackage{pifont}
\usepackage{amssymb}
\usepackage{enumitem}
\title{HyperReCo: Retrieving and Connecting Evidence with Hypergraph Neural Networks for LLM Multi-hop Reasoning}

\usepackage{etoolbox}

\makeatletter
\patchcmd{\@maketitle}
  {Anonymous authors\\Paper under double-blind review}
  {\@author}
  {}
  {\PackageError{arxiv-version}
    {Failed to patch the anonymous author block}
    {Check the definition of \string\@maketitle\space in your ICLR style file.}}
\makeatother

\renewcommand{\iclrruler}[1]{}

\author{
Zicheng Zhao\textsuperscript{1, 6},
Linhao Luo\textsuperscript{2},
Junnan Dong\textsuperscript{4},
Haoran Luo\textsuperscript{5},
Xiaoli Li\textsuperscript{6}, \\[1pt]
\textbf{Shirui Pan\textsuperscript{3*},
Chen Gong\textsuperscript{1*}}
\\[2pt]
\textsuperscript{1}Shanghai Jiao Tong University,
\textsuperscript{2}Monash University,
\textsuperscript{3}Griffith University,
\textsuperscript{4}Tencent Youtu Lab,  \\
\textsuperscript{5}Nanyang Technological University,
\textsuperscript{6}Singapore University of Technology and Design
\\[1pt]
\texttt{zicheng.zhao99@gmail.com, chen.gong@sjtu.edu.cn}
\\[1pt]
\textsuperscript{*}Corresponding author.
}

\newcommand{\ourmethod}{\texttt{HyperReCo}\xspace}

\begin{document}

\maketitle

\fancyhead{}

\begin{abstract}
Large language models (LLMs) have shown strong capabilities, with retrieval-augmented generation (RAG) supporting complex multi-hop reasoning by retrieving evidence distributed across documents. Graph-based approaches exploit connections among evidence, and hypergraph-based retrieval further preserves higher-order entity associations within documents and connects documents through shared entities.
However, existing hypergraph retrievers often rely on predefined structural expansion or diffusion, which may miss query-dependent interactions needed to identify relevant evidence. They also leave connections among retrieved evidence implicit, requiring LLMs to reconstruct these connections before reasoning. Therefore, we propose \ourmethod, a framework for retrieving and connecting evidence with a hypergraph neural network (HyperGNN). We represent each document as a hyperedge over its extracted entities, with shared entities connecting the hyperedges. Through hypergraph message passing with joint supervision over documents and entities, the HyperGNN learns query-dependent interactions to retrieve complementary evidence. We further introduce Gradient-Guided Hyper-Path Decoding (GGHD), which uses gradient attribution to interpret the learned interactions and translate them into explicit hyper-paths that help LLMs combine complementary facts for multi-hop reasoning. 
Experiments on six benchmarks show that \ourmethod achieves the best retrieval performance among the compared methods on all three multi-hop QA datasets, together with strong downstream QA performance. Case studies and further analyses demonstrate the utility of decoded hyper-paths for connecting retrieved evidence.


\end{abstract}

\section{Introduction}
\label{sec:intro}

Large language models (LLMs) have demonstrated strong capabilities~\citep{achiam2023gpt}. However, solving complex tasks often requires multi-hop reasoning over evidence from multiple sources~\citep{yang2018hotpotqa,ho2020constructing,trivedi2022musique}. This motivates the need for methods that can effectively \emph{retrieve} and \emph{connect} evidence across different documents to support LLM reasoning~\citep{trivedi2023interleaving}. Retrieval-augmented generation (RAG) has been proposed to address this challenge by retrieving relevant documents that provide evidence for LLM reasoning~\citep{lewis2020retrieval}. Recently, GraphRAG~\citep{edge2024local} has been introduced to extend RAG by organizing knowledge into graph structures and capturing the evidence connections via graph search, relevance propagation, or learned message passing for improved reasoning~\citep{jimenez2024hipporag,luo2025gfm,yan2026questgnn}.


Despite the advances, GraphRAG methods are often criticized for the cost overhead of extracting relational triples to construct knowledge graph (KG) indexes~\citep{edge2024local}. More importantly, KG triples only encode binary relations, making them insufficient for modeling the n-ary relations among multiple entities that are widespread in real-world domain knowledge~\citep{3060621.3060802, huang2024link,huang2025hyper}. Recently, hypergraph-based retrieval methods have been proposed to address this limitation by representing evidence as hyperedges over sets of entities~\citep{luo2026hypergraphrag,wang2026cross,feng2026hyper}. Hypergraphs can capture higher-order connections across evidence through sharing entities without explicit relation extraction, which makes them a promising alternative for efficient evidence retrieval. 

However, existing hypergraph retrievers often rely on predefined structural expansion or graph diffusion mechanisms~\citep{song2026ehrag,feng2026hyper}. Although hyperedges can encode complex n-ary relations among multiple entities, retrieving useful evidence requires identifying how these entities jointly relate to the query. Structural expansion or graph diffusion alone may not adequately capture such query-dependent interactions, making it difficult to distinguish relevant evidence from connected but irrelevant content~\citep{luo2026hypergraphrag,wang2026cross}. Consequently, these retrievers may miss complementary evidence needed for multi-hop reasoning.

\begin{wrapfigure}{r}{0.55\textwidth}
    \centering
    \vspace{-1.1\baselineskip}
    \includegraphics[width=0.48\textwidth]{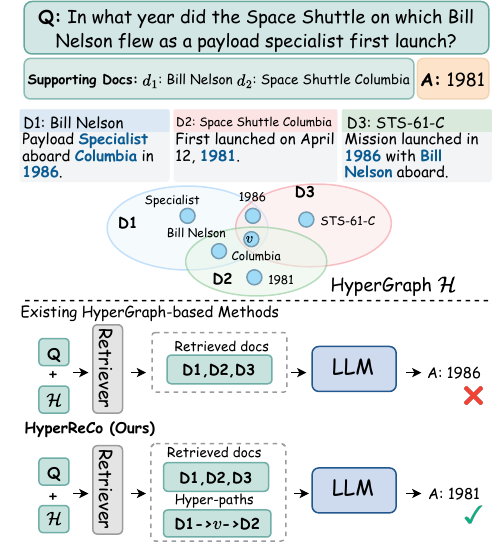}
    \caption{Illustration of the evidence routing gap. \ourmethod augments the retrieved evidence with a decoded hyper-path linking retrieved documents.}
    \vspace{-1.2\baselineskip}
    \label{fig:intro}
\end{wrapfigure}

Moreover, existing hypergraph retrievers often leave the connection among retrieved evidence implicit~\citep{wang2026cross,feng2026hyper}. Exploiting hypergraph connectivity during retrieval does not by itself provide LLMs with an explicit representation of which documents contain complementary facts and how they are linked. Therefore, LLMs must reconstruct these connections before reasoning. For example,  Figure~\ref{fig:intro} shows the question: ``In what year did the Space Shuttle on which Bill Nelson flew as a payload specialist first launch?'' Answering this question requires linking the document about \emph{Bill Nelson}'s shuttle mission to the document about \emph{Space Shuttle Columbia}'s first launch through the shared entity \emph{Columbia}. Explicitly representing this connection can help LLMs distinguish the shuttle's first launch from Nelson's later mission and facilitate reasoning. Addressing these challenges requires \emph{capturing query-dependent interactions to retrieve relevant evidence} and \emph{explicitly representing connections among the retrieved evidence} to guide LLM reasoning.

To address these challenges, we propose \ourmethod, a framework for retrieving and connecting evidence with a hypergraph neural network (HyperGNN) to enhance LLM multi-hop reasoning. We first extract entities from the documents and represent each document as a hyperedge connecting its entity nodes. These hyperedges are linked through shared entities to form a hypergraph. On this hypergraph, we propose a \textbf{HyperGNN} to capture higher-order interactions through hypergraph message passing, which aggregates information within hyperedges and propagates it across hyperedges through shared entities. Through joint supervision of document and entity relevance, the model learns to capture query-dependent interactions that guide the retrieval of complementary evidence. We then introduce \textbf{Gradient-Guided Hyper-Path Decoding (GGHD)} to interpret the query-dependent interactions learned by the HyperGNN and translate them into explicit hyper-paths connecting the retrieved evidence. Building on gradient-based graph interpretation~\citep{zhu2021neural,luo2025gfm}, GGHD performs gradient attribution on the trained HyperGNN to identify how messages passed through shared entities influence its document relevance predictions. It uses these attributions to derive connection costs and combines them with document-dependent connectivity penalties in a query-conditioned prize-collecting Steiner forest (Q-PCSF) objective~\citep{ahmadi20252}. The resulting forest is serialized into grouped hyper-paths that make the connections through shared entities explicit, helping the LLM connect complementary facts across documents for multi-hop reasoning.

In experiments, \ourmethod achieves state-of-the-art performance across three widely-used QA datasets, which demonstrates the effectiveness and efficiency in multi-hop reasoning. Meanwhile, \ourmethod also generalizes well to domain-specific QA datasets spanning medical, novel, and computer science domains without domain-specific training. Ablation studies assess the contributions of the core components, while quantitative and qualitative analysis show that the decoded hyper-paths provide explicit evidence connections that improve downstream question answering.


\section{Related Work}
\label{sec:relatedwork}


\noindent\textbf{Graph Retrieval-Augmented Generation.}
GraphRAG augments semantic retrieval with structured knowledge to support evidence retrieval and reasoning. GraphRAG~\citep{edge2024local} combines entity graphs with community summaries, LightRAG~\citep{guo2024lightrag} integrates graph indexing with dual-level retrieval, and HippoRAG~\citep{jimenez2024hipporag} propagates query relevance over extracted knowledge graphs using personalized PageRank. Learning-based approaches further model query-dependent structural relevance. GFM-RAG~\citep{luo2025gfm} learns transferable message passing over knowledge graphs, and Quest-GNN~\citep{yan2026questgnn} performs question-guided propagation over multi-level graphs. \ourmethod{} follows this learning-based direction but operates over evidence-bearing hyperedges, avoiding explicit relation extraction while learning higher-order structural relevance.

\noindent\textbf{Hypergraph Reasoning.}
Hypergraphs naturally represent higher-order associations among multiple entities and have recently been incorporated into retrieval-augmented generation. HyperGraphRAG~\citep{luo2026hypergraphrag} represents n-ary facts as hyperedges, while Hyper-RAG~\citep{feng2026hyper} combines semantic retrieval with expansion over higher-order correlations. HGRAG~\citep{wang2026cross} grounds hyperedges in passages and performs hypergraph diffusion, and EHRAG~\citep{song2026ehrag} further introduces semantic hyperedges constructed from entity clusters. HyperRAG~\citep{lien2026hyperrag} instead learns a plausibility scorer for query-conditioned traversal. These methods demonstrate the value of higher-order structure, but differ in how query relevance is propagated or learned. In contrast, \ourmethod{} trains a query-conditioned HyperGNN with joint document and entity supervision, enabling structural relevance to be learned directly over overlapping evidence hyperedges.

\noindent\textbf{Path Interpretability.}
Paths provide an explicit view of how structural information contributes to graph predictions. NBFNet~\citep{zhu2021neural} uses gradient-based attribution to identify influential reasoning paths, and GFM-RAG~\citep{luo2025gfm} similarly analyzes learned graph predictions through gradients. Other approaches directly use paths as retrieval or generation context: PathRAG~\citep{chen2026pathrag} retrieves relational paths for LLM prompting, while G-retriever~\citep{he2024g} formulates subgraph selection as prize-collecting Steiner tree optimization. Our GGHD connects these two directions: it uses message sensitivities from the trained HyperGNN to identify query-relevant entity-mediated connections and organizes them with a query-conditioned forest objective. The resulting grouped hyper-paths translate learned structural relevance into explicit evidence connections for downstream LLM reasoning.
\section{Preliminary}\label{sec:preliminary}

\noindent\textbf{Multi-hop Reasoning.}
Multi-hop reasoning over documents aims to answer a question by combining complementary facts distributed across multiple documents and using the connections among them. Formally, given a question $q$ and a corpus $\mathcal{D}=\{d_1,\ldots,d_N\}$, we seek to design a retriever $f_{\mathrm{ret}}$ with parameters $\theta$ that selects relevant documents $\mathcal{D}_q=f_{\mathrm{ret}}(q,\mathcal{D},\theta)$ and captures their connections $d_1\xrightarrow{r_1} d_2\xrightarrow{r_2}\cdots\xrightarrow{r_{h-1}} d_h$ that reveal how their facts jointly support an answer $a$. 



\noindent\textbf{Hypergraph.}
We represent the corpus as a hypergraph $\mathcal{H}=(\mathcal{V},\mathcal{E})$, with entities as nodes and documents as hyperedges. Following prior work~\citep{wang2026cross}, we use LLM-based named entity recognition (NER) to extract entity mentions, and let $\mathcal{V}$ denote the resulting set of canonical entities. Each document $d\in\mathcal{D}$ induces a hyperedge as
\begin{equation}
    e_d=\{v\in\mathcal{V}\mid v\text{ has a mention in }d\}.
\end{equation}
The hyperedge family $\mathcal{E}=(e_d)_{d\in\mathcal{D}}$ is indexed by documents, retaining a separate hyperedge for each document even when their entity sets coincide. An entity mentioned in multiple documents belongs to each corresponding hyperedge and connects them. This representation preserves within-document entity associations and cross-document connections without extracting relational triples.

\noindent\textbf{Hyper-Path.}
For distinct documents $d_0,\ldots,d_h$ with $h\geq 1$, we define a \emph{hyper-path} as a sequence of their hyperedges connected through nonempty shared entity sets:
\begin{equation}
    \mathcal{R}=e_{d_0}\xrightarrow{\widehat{\mathcal{B}}_{0,1}}e_{d_1}\xrightarrow{\widehat{\mathcal{B}}_{1,2}}\cdots\xrightarrow{\widehat{\mathcal{B}}_{h-1,h}}e_{d_h},
    \label{eq:preliminary:hyperpath}
\end{equation}
where $\varnothing\neq\widehat{\mathcal{B}}_{t-1,t}\subseteq\mathcal{B}_{t-1,t}=e_{d_{t-1}}\cap e_{d_t}$ for $t=1,\ldots,h$. Here, $\mathcal{B}_{t-1,t}$ contains all entities shared by the two documents, while $\widehat{\mathcal{B}}_{t-1,t}$ contains those used to connect them along the path. The path spans $h$ document-to-document transitions and makes their entity-mediated connections explicit, providing structural guidance for reasoning over the source documents.

\section{Approach}
\label{sec:approach}

\begin{figure*}[t]
    \centering
    \includegraphics[width=\textwidth]{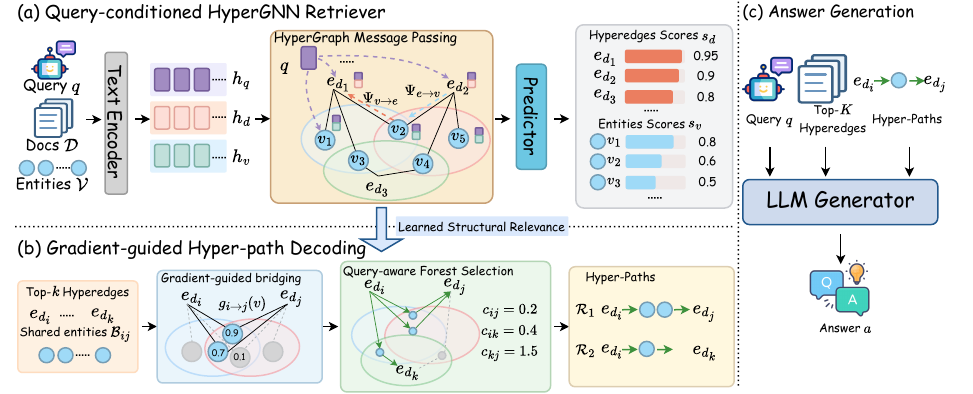}
    \caption{Overview of \ourmethod. 
    (a) The query-conditioned HyperGNN learns structural relevance over hypergraph for evidence retrieval. 
    (b) Gradient-Guided Hyper-Path Decoding (GGHD) uses gradient attribution and Q-PCSF to translate the learned structural relevance into explicit grouped hyper-paths. 
    (c) The retrieved documents and decoded hyper-paths are provided to the LLM for answer generation.}
    \label{fig:framework}
    \vspace{-.4cm}
\end{figure*}

Building on the hypergraph constructed in Section~\ref{sec:preliminary}, \ourmethod combines two complementary modules to \emph{retrieve relevant evidence} and \emph{make its connections explicit} for LLM reasoning. For evidence retrieval, we propose a \textbf{HyperGNN} retriever (Figure~\ref{fig:framework}(a)) to capture query-dependent interactions and retrieve complementary evidence. To connect the retrieved evidence, we introduce \textbf{Gradient-Guided Hyper-Path Decoding (GGHD)} (Figure~\ref{fig:framework}(b)), which uses gradient attribution to identify the query-dependent relevance captured by the HyperGNN and translate it into explicit hyper-paths connecting the retrieved documents. These paths guide the LLM in combining complementary facts across documents.

\subsection{HyperGraph Neural Network Retriever}
\label{sec:approach:encoding}



Although hypergraphs model complex higher-order connections, their overlapping entity associations introduce enormous candidate evidence links, making it difficult to identify query-relevant interactions from structure alone~\citep{song2026ehrag,wang2026cross}. Graph neural networks can learn expressive structural representations on hypergraphs~\citep{huang2024link,huang2025hyper} and have also been applied to evidence retrieval~\citep{luo2025gfm}. We therefore propose a HyperGNN retriever that learns query-dependent interactions through alternating hypergraph message passing to retrieve complementary evidence.

\noindent\textbf{Query-dependent Representation Initialization.} We first adopt a frozen text encoder to encode the query, documents, and extracted entities into embeddings $x_q$, $x_d$, and $x_v$, respectively. We then project $x_q$ to obtain the query representation $h_q$, and inject this query signal into the document and entity representations to initialize the corresponding hyperedge and entity states, denoted by $h_{e_d}^{(0)}$ and $h_v^{(0)}$. Detailed initialization is provided in Appendix~\ref{sec:app:hypergnn}.

\noindent\textbf{Hypergraph Message Passing.} At each layer $\ell$, the HyperGNN alternates between hyperedge-to-entity and entity-to-hyperedge propagation. We first aggregate information from the hyperedges containing entity $v$ to update its representation as:
\begin{equation}
    h_v^{(\ell+1)}=\Phi_V^{(\ell)}\left(h_v^{(\ell)},\operatorname{AGG}_{e_d:\,v\in e_d}\Psi_{E\rightarrow V}^{(\ell)}\bigl(h_{e_d}^{(\ell)},h_v^{(\ell)},h_q\bigr)\right),
\label{eq:approach:edge-to-entity}
\end{equation}
where $\Psi_{E\rightarrow V}^{(\ell)}$ computes query-conditioned messages from hyperedges to entities, $\operatorname{AGG}$ performs attention-based aggregation, and $\Phi_V^{(\ell)}$ updates the entity state. The updated entity representations are then propagated back to their associated hyperedges:
\begin{equation}
    h_{e_d}^{(\ell+1)}=\Phi_E^{(\ell)}\left(h_{e_d}^{(\ell)},\operatorname{AGG}_{v\in e_d}\Psi_{V\rightarrow E}^{(\ell)}\bigl(h_v^{(\ell+1)},h_{e_d}^{(\ell)},h_q\bigr)\right),
\label{eq:approach:entity-to-edge}
\end{equation}
where $\Psi_{V\rightarrow E}^{(\ell)}$ and $\Phi_E^{(\ell)}$ denote the corresponding message and update functions. By repeatedly exchanging information between entities and hyperedges, the HyperGNN learns how shared entities contribute differently across evidence units under the current query, thereby capturing query-conditioned structural relevance. Detailed message functions and attention mechanisms are provided in Appendix~\ref{sec:app:hypergnn-mp}.

\noindent\textbf{Relevance Scoring.} After $L$ propagation layers, we predict relevance for both hyperedges and entities:
\begin{equation}
    s_d=f_{\mathrm{doc}}\bigl([h_{e_d}^{(L)};h_q]\bigr),
    s_v=f_{\mathrm{ent}}\bigl([h_v^{(L)};h_q]\bigr),
\end{equation}
where $s_d$ denotes the relevance logit associated with hyperedge $e_d$, and $s_v$ denotes the relevance logit of entity $v$. Joint supervision at the hyperedge and entity levels encourages the HyperGNN to learn structural relevance from both evidence and its constituent entities. We rank the hyperedges by $s_d$ and return the source documents corresponding to the top-$K$ hyperedges as $\mathcal{D}_q$.

\subsection{Gradient-Guided Hyper-Path Decoding}
\label{sec:approach:gghd}

The predicted relevance scores identify important evidence units, but do not explicitly reveal how the retrieved evidence should be connected. We therefore introduce \textbf{Gradient-Guided Hyper-Path Decoding (GGHD)} to expose the structural relevance learned by the HyperGNN as explicit evidence connections. Inspired by gradient-based path interpretation~\citep{zhu2021neural,luo2025gfm}, GGHD attributes document predictions to entity-mediated messages and organizes the resulting connections with a query-conditioned prize-collecting Steiner forest (Q-PCSF).

\noindent\textbf{Gradient-guided connections.}
Let $\mathcal{D}_q$ denote the set of documents retrieved for query $q$, and let $e_d$ be the document-induced hyperedge associated with document $d$. For two distinct retrieved documents $d_i,d_j\in\mathcal{D}_q$, their shared-entity set is $\mathcal{B}_{ij}=e_{d_i}\cap e_{d_j}$. To measure how a shared entity $v\in\mathcal{B}_{ij}$ contributes to the relevance prediction of $d_j$, we place independent scalar gates $\gamma_{e_{d_i}\rightarrow v}^{(\ell)}$ and $\gamma_{v\rightarrow e_{d_j}}^{(\ell)}$ on the attention-normalized hyperedge-to-entity and entity-to-hyperedge messages at propagation layer $\ell$, respectively. The model parameters are frozen during attribution, and all gates are evaluated at one so that the original forward prediction is unchanged. We define the directional bridge attribution as
\begin{equation}
    \small
    g_{i\rightarrow j}^{(\ell)}(v)
    =
    \frac{1}{2}
    \left[
        \frac{\partial s_{d_j}}{\partial \gamma_{e_{d_i}\rightarrow v}^{(\ell)}}
        +
        \frac{\partial s_{d_j}}{\partial \gamma_{v\rightarrow e_{d_j}}^{(\ell)}}
    \right]_{\boldsymbol{\gamma}=\mathbf{1}},
    \label{eq:approach:bridge-gradient}
\end{equation}
where $s_{d_j}$ is the relevance logit predicted by the HyperGNN for target document $d_j$. Therefore, $g_{i\rightarrow j}^{(\ell)}(v)$ measures the local first-order sensitivity of the target prediction to the two messages associated with the entity-mediated connection $e_{d_i}\rightarrow v\rightarrow e_{d_j}$. Since two documents may share many incidental entities, for each direction and propagation layer we retain only the Top-$M$ shared entities with positive attribution, denoted by $\widehat{\mathcal{B}}_{i\rightarrow j}^{(\ell)}\subseteq\mathcal{B}_{ij}$.

\noindent\textbf{Gradient-derived connection costs.}
We next convert the directional bridge attributions into an undirected connection score between each pair of retrieved documents. Let
\begin{equation}
    w_j=\frac{\exp(s_{d_j})}{\sum_{d\in\mathcal{D}_q}\exp(s_d)}, \qquad
    b_{i\rightarrow j}^{(\ell)}
    =
    w_j
    \max_{v\in\widehat{\mathcal{B}}_{i\rightarrow j}^{(\ell)}}
    g_{i\rightarrow j}^{(\ell)}(v),
    \label{eq:approach:directional-strength}
\end{equation}
where $w_j$ is the normalized relevance of target document $d_j$ within the retrieved set and $b_{i\rightarrow j}^{(\ell)}$ measures the relevance-weighted strength of the corresponding directional connection at layer $\ell$. Because the same document pair may interact in either direction and at multiple propagation layers, we aggregate them by
\begin{equation}
    \small
    b_{ij}
    =
    \max_{\ell}
    \left\{
        b_{i\rightarrow j}^{(\ell)},
        b_{j\rightarrow i}^{(\ell)}
    \right\},
    \qquad
    c_{ij}
    =
    \log\left(1+\frac{\mu_q}{b_{ij}}\right),
    \label{eq:approach:connection-cost}
\end{equation}
where $b_{ij}$ is the final connection strength and $\mu_q$ is the median positive connection strength among candidate document pairs for query $q$. The resulting $c_{ij}$ serves as the connection cost: document pairs whose relevance predictions are more strongly supported by shared-entity interactions receive lower costs. We retain only pairs with $b_{ij}>0$, yielding the candidate connection set $E_q$ over the retrieved documents.

\noindent\textbf{Query-conditioned forest decoding.}
Gradient-derived costs characterize individual connections, whereas evidence routing requires selecting connections jointly to organize the retrieved evidence. Inspired by prize-collecting graph retrieval~\citep{he2024g}, we use a query-conditioned prize-collecting Steiner forest (Q-PCSF) objective to balance connection costs against relevance-dependent connectivity demands. Let $G_q=(\mathcal{D}_q,E_q)$ denote the candidate graph, and select the highest-scoring retrieved document $d_a\in\operatorname*{arg\,max}_{d_i\in\mathcal{D}_q}s_{d_i}$ as the anchor.
For every other retrieved document $d_j$, we introduce an anchor-to-document connectivity demand with penalty as
\begin{equation}
    \pi_{aj}
    =
    \lambda\sqrt{p_{d_a}p_{d_j}},
    \qquad
    p_d=\sigma(s_d),
    \label{eq:approach:connectivity-demand}
\end{equation}
where $p_d$ is the predicted relevance probability and $\lambda>0$ controls the trade-off between selecting additional connections and leaving relevant documents disconnected. Higher-relevance documents therefore incur larger penalties when they are not connected to the anchor.

We select the forest edge set by minimizing
\begin{equation}
    \small
    F_q^\star
    \in
    \operatorname*{arg\,min}_{F\in\mathfrak{F}_q}
    \left\{
        \sum_{\{d_i,d_j\}\in F} c_{ij}
        +
        \sum_{d_j\in\mathcal{D}_q\setminus\{d_a\}}
        \pi_{aj}
        \mathbb{1}[d_a\not\leftrightarrow_F d_j]
    \right\},
    \label{eq:approach:qpcsf}
\end{equation}
where $\mathfrak{F}_q$ is the family of edge sets $F\subseteq E_q$ for which $(\mathcal{D}_q,F)$ forms a forest. The indicator $\mathbb{1}[d_a\not\leftrightarrow_F d_j]$ equals one when the anchor and document $d_j$ belong to different connected components. The first term favors strongly attributed connections through their lower costs, while the second penalizes unsatisfied connectivity demands. A demand can be satisfied through intermediate retrieved documents, allowing multiple connections to jointly support an evidence route. The objective also permits demands to remain unsatisfied when the additional connection costs outweigh the penalties avoided. The detailed optimization procedure is provided in Appendix~\ref{sec:app:qpcsf}.

\subsection{Optimization and Answer Generation}
\label{sec:approach:optimization_generation}

Given the supporting-document set $\mathcal{S}_q$, we supervise hyperedge relevance using document labels and derive weak entity supervision from entities appearing in the supporting documents~\citep{luo2025gfm}. Together with a KL-based semantic distillation objective, the retriever is optimized as
\begin{equation}
    \theta^\star
    =
    \operatorname*{arg\,min}_{\theta}
    \mathbb{E}_{q\sim\mathcal{Q}_{\mathrm{train}}}
    \left[
    \mathcal{L}_{\mathrm{ret}}(\theta;q)
    +
    \lambda_{\mathrm{KL}}\mathcal{L}_{\mathrm{KL}}(\theta;q)
    \right],
\label{eq:approach:objective}
\end{equation}
where $\lambda_{\mathrm{KL}}$ balances supervised relevance learning and semantic distillation. Detailed definitions of the supervision signals and loss terms are provided in Appendix~\ref{sec:app:optimization}.

After training, GGHD decodes the learned structural relevance without updating $\theta^\star$. The resulting forest $F_q^\star$ is serialized into hyper-paths $\mathcal{R}_q$ and provided together with the retrieved documents $\mathcal{D}_q$ for answer generation, which can be formulated as
\begin{equation}
    a=f_{\mathrm{gen}}(q,\mathcal{D}_q,\mathcal{R}_q).
\label{eq:approach:answer}
\end{equation}

The hyper-paths provide explicit guidance for composing evidence across documents, while factual claims remain grounded in the retrieved content. Serialization and prompt details are provided in Appendix~\ref{sec:app:prompt}.

\section{Experiment} \label{sec:exp}

In this section, we attempt to answer the following research questions:
\textbf{RQ1:} How does \ourmethod perform in multi-hop evidence retrieval and question answering compared with existing methods? 
\textbf{RQ2:} Can \ourmethod effectively generalize across different domains?
\textbf{RQ3:} How do the key components of \ourmethod contribute to the overall performance? 
\textbf{RQ4:} How do the hyper-paths derived from \ourmethod contribute to the downstream task?
\textbf{RQ5:} How efficient is \ourmethod in terms of index construction and inference?

\begin{table}[]

\centering
\caption{QA reasoning performance comparison. GPT-4o-mini is used as the LLM for reasoning. The best result is in bold.}
\label{tab:qa}
\vspace{-0.3cm}
\resizebox{0.95\columnwidth}{!}{%
\begin{tabular}{@{}l|cc|cc|cc|c|c|c@{}}
\toprule
Method
& \multicolumn{2}{c|}{HotpotQA}
& \multicolumn{2}{c|}{MuSiQue}
& \multicolumn{2}{c|}{2Wiki}
& \shortstack{G-bench\\(Novel)}
& \shortstack{G-bench\\(Medical)}
& \shortstack{G-bench\\(CS)} \\
\cmidrule(l){2-10}
& EM   & F1
& EM   & F1
& EM   & F1
& ACC  & ACC
& ACC \\
\midrule

\multicolumn{10}{c}{\cellcolor[HTML]{C0C0C0}Non-structure Methods} \\
None (GPT-4o-mini)~\citep{gpt4o}          & 28.6 & 41.0 & 11.2 & 36.3 & 30.2 & 36.3 & 51.4       & 67.1       & 70.7 \\
BM25 \citep{robertson1994some}            & 52.0 & 63.4 & 20.3 & 28.8 & 47.9 & 51.2 & 56.5       & 68.7       & 71.7 \\
ColBERTv2 \citep{santhanam2022colbertv2}  & 43.4 & 57.7 & 15.5 & 26.4 & 33.4 & 43.3 & 56.2       & 71.8       & 71.9 \\
Qwen3-Emb (8B) \citep{zhang2025qwen3}     & 53.4 & 67.6 & 31.9 & 44.1 & 57.2 & 63.2 & 56.2       & 70.4       & 73.5 \\
NV-Embed-v2 (7B) \citep{lee2025nv}        & 56.9 & 70.9 & 31.8 & 45.5 & 57.6 & 64.3 & \textbf{65.3}       & 77.7      & 74.2 \\
\midrule

\multicolumn{10}{c}{\cellcolor[HTML]{C0C0C0}Graph-enhanced Methods} \\
RAPTOR \citep{sarthiraptor}               & 50.6 & 64.7 & 27.7 & 39.2 & 39.7 & 48.4 & 43.2       & 57.1       & 73.6 \\
GraphRAG (MS) \citep{edge2024local}       & 51.4 & 67.6 & 27.0 & 42.0 & 34.7 & 61.0 & 50.9       & 45.2       & 72.5 \\
LightRAG \citep{guo2024lightrag}          &  9.9 & 20.2 &  2.0 &  9.3 &  2.5 & 12.1 & 45.1       & 63.9       & 71.2 \\
HippoRAG \citep{jimenez2024hipporag}      & 46.3 & 60.0 & 24.0 & 35.9 & 59.4 & 67.3 & 44.8       & 59.1       & 72.6 \\
HippoRAG 2 \citep{gutierrez2025ragmemory} & 56.3 & {71.1} & {35.0} & {49.3} & 60.5 & 69.7 & 56.5       & 64.9       & -    \\
GFM-RAG \citep{luo2025gfm}                & 56.2 & 69.5 & 30.2 & 49.2 & {69.8} & {77.7} & 58.6       & 72.2       & 72.1 \\
G-retriever \citep{he2024g}               & 41.4 & 53.4 & 23.6 & 34.3 & 33.5 & 39.6 & -          & -          & 69.8 \\
Quest-GNN \citep{yan2026questgnn}         & 54.9 & 68.3 & 28.5 & 40.1  & 51.5 & 57.7 & -  & -  & - \\
EHRAG \citep{song2026ehrag}               & 54.5 & 68.9 & 34.8 & 48.5  & 65.3 & 73.1 & 64.8 & 75.9  & 73.9 \\
HGRAG \citep{wang2026cross}               & 55.9 & 70.5 & 36.2 & 48.6  & 68.2 & 76.3 & 64.9 & 76.1  & \textbf{74.5} \\

\midrule

\ourmethod                                 & \textbf{60.7} & \textbf{74.0} & \textbf{39.9} & \textbf{51.5} & \textbf{72.2} & \textbf{78.7} & 65.2 & \textbf{78.4} & 73.9 \\
\bottomrule
\end{tabular}%
}
\vspace{-.4cm}
\end{table}

\subsection{Experimental Setup} \label{sec:exp:exp-set}
\textbf{Datasets.} We first evaluate the effectiveness of \ourmethod on three widely-used multi-hop QA datasets, consisting of HotpotQA~\citep{yang2018hotpotqa}, MuSiQue~\citep{trivedi2022musique}, and 2WikiMultiHopQA (2Wiki)~\citep{ho2020constructing}, following the settings used in ~\citep{luo2025gfm, wang2026cross} for a fair comparison. To assess the generalization of \ourmethod across different domains, we further evaluate \ourmethod on three GraphRAG benchmark datasets: G-bench (Novel)~\citep{xiang2025use}, G-bench (Medical)~\citep{xiang2025use}, and G-bench (CS)~\citep{xiao2025graphrag}. Detailed statistics of both the training and evaluation datasets are provided in Appendix~\ref{app:dataset_statistics}.  

\noindent \textbf{Baseline methods.} We compare \ourmethod with two groups of baseline methods: (1) \textit{Non-structure Methods}: BM25~\citep{robertson1994some}, ColBERTv2~\citep{santhanam2022colbertv2}, Qwen3-Embedding-8B (Qwen3-Emb)~\citep{zhang2025qwen3}, and NV-Embed-v2 (7B)~\citep{lee2025nv}; (2) \textit{Graph-enhanced methods}: RAPTOR~\citep{sarthiraptor}, GraphRAG~\citep{edge2024local}, LightRAG~\citep{guo2024lightrag}, HippoRAG~\citep{jimenez2024hipporag}, HippoRAG 2~\citep{gutierrez2025ragmemory}, GFM-RAG~\citep{luo2025gfm}, G-retriever~\citep{he2024g}, Quest-GNN~\citep{yan2026questgnn}, EHRAG~\citep{song2026ehrag}, and HGRAG~\citep{wang2026cross}. For a fair comparison, we re-run EHRAG and HGRAG using the same LLM for graph construction and retrain Quest-GNN on the same training datasets used in our experiments. Detailed descriptions of these baseline methods are provided in Appendix~\ref{sec:app:baseline}.

\noindent \textbf{Metrics.} We evaluate \ourmethod from two perspectives: downstream QA performance and retrieval performance. For multi-hop QA, we report Exact Match (EM) and F1 following prior work~\citep{jimenez2024hipporag,luo2025gfm}, while for the G-bench datasets, we report Accuracy (ACC) following their original evaluation protocols~\citep{xiang2025use,xiao2025graphrag}. For retrieval performance, we use document recall@2 (R@2) and recall@5 (R@5) on multi-hop QA datasets, and evidence recall (Recall) on the G-bench datasets~\citep{xiang2025use}.

\noindent \textbf{Implementation Details.} We train \ourmethod using large-scale datasets curated from GFM-RAG~\citep{luo2025gfm}, which consists of 6,000 query samples and 53,573 documents. GPT-4o-mini is consistently used for QA inference across all methods. More training and implementation details are provided in Appendix~\ref{sec:app:implementation}.

\subsection{Main Results (RQ1 \& RQ2)}

\noindent\textbf{Multi-hop QA and Retrieval.}
As shown in~\Cref{tab:qa,tab:retrieval}, \ourmethod achieves the strongest performance on all three multi-hop QA benchmarks in both downstream QA and evidence retrieval. In particular, the consistent gains in R@2 and R@5 indicate that the query-conditioned HyperGNN better prioritizes complementary evidence beyond semantic similarity alone. These retrieval improvements are accompanied by higher EM and F1 scores, suggesting that combining learned structural relevance with explicit hyper-path guidance benefits downstream multi-hop reasoning. Compared with retrievers based on predefined search or propagation, \ourmethod learns query-dependent interactions directly over hypergraphs and further exposes the resulting evidence connections to the LLM through GGHD.

\noindent\textbf{Cross-domain Generalization.}
We further evaluate \ourmethod on G-bench datasets covering novel, medical, and computer science domains to examine its generalization ability. As shown in~\Cref{tab:qa,tab:retrieval}, \ourmethod remains competitive with strong semantic and graph-enhanced baseline methods without domain-specific training, and achieves the best QA performance on the medical benchmark. Notably, its QA performance can remain strong even when evidence recall is comparable to or slightly below the strongest retrievers, indicating that downstream reasoning benefits not only from relevant evidence, but also from explicit evidence connections.

\begin{table}[]
\centering
\caption{Retrieval performance comparison. Recall@$k$ (R@$k$) is used for multi-hop QA datasets, and evidence recall (Recall) is used for G-bench~\citep{xiang2025use}. The best result is in bold.}
\label{tab:retrieval}
\vspace{-0.3cm}
\resizebox{.95\columnwidth}{!}{%
\begin{tabular}{@{}l|cc|cc|cc|c|c@{}}
\toprule
Method
& \multicolumn{2}{c|}{HotpotQA}
& \multicolumn{2}{c|}{MuSiQue}
& \multicolumn{2}{c|}{2Wiki}
& \shortstack{G-bench\\(Novel)}
& \shortstack{G-bench\\(Medical)} \\
\cmidrule(l){2-9}
& R@2 & R@5
& R@2 & R@5
& R@2 & R@5
& Recall & Recall \\
\midrule

\multicolumn{9}{c}{\cellcolor[HTML]{C0C0C0}Non-structure Methods} \\
BM25 \citep{robertson1994some}               & 55.4 & 72.2 & 32.3 & 41.2 & 51.8 & 61.9 & 82.1       & 87.9 \\
ColBERTv2 \citep{santhanam2022colbertv2}     & 64.7 & 79.3 & 37.9 & 49.2 & 59.2 & 68.2 & 82.4       & 89.5 \\
Qwen3-Emb (8B) \citep{zhang2025qwen3}        & 74.1 & 88.8 & 46.8 & 62.1 & 66.2 & 74.1 & 82.6       & 92.7 \\
NV-Embed-v2 (7B) \citep{lee2025nv}           & {84.1} & 94.4 & 52.7 & 69.5 & 69.1 & 76.5 & \textbf{92.5}       & \textbf{96.5} \\
\midrule

\multicolumn{9}{c}{\cellcolor[HTML]{C0C0C0}Graph-enhanced Methods} \\
RAPTOR \citep{sarthiraptor}                  & 58.1 & 71.2 & 35.7 & 45.3 & 46.3 & 53.8 & 66.1       & 84.2 \\
GraphRAG (MS) \citep{edge2024local}          & 58.3 & 76.6 & 35.4 & 49.3 & 61.6 & 77.3 & 67.4       & 56.4 \\
LightRAG \citep{guo2024lightrag}             & 38.8 & 54.7 & 24.8 & 34.7 & 45.1 & 59.1 & 79.6       & 82.6 \\
HippoRAG \citep{jimenez2024hipporag}         & 60.1 & 78.5 & 41.2 & 53.2 & 68.4 & 87.0 & 81.2       & 84.0 \\
HippoRAG 2 \citep{gutierrez2025ragmemory}    & 80.5 & {95.7} & 53.5 & {74.2} & {75.2} & {90.5} & 66.2       & 73.6 \\
G-retriever \citep{he2024g}                  & 51.8 & 63.6 & 35.6 & 43.5 & 60.9 & 66.5 & -          & -    \\
GFM-RAG \citep{luo2025gfm}                   & 75.6 & 89.6 & 43.5 & 57.6 & 79.1 & 92.4 & 75.9  & 82.2 \\
Quest-GNN \citep{yan2026questgnn}          & 76.7 & 89.1 & 38.7 & 54.0 & 61.1 & 71.2 & -  & - \\
EHRAG \citep{song2026ehrag}               & 76.6 & 92.7 & 53.0 & 70.7  & 69.4 & 85.6 & 92.3 & 93.0\\
HGRAG \citep{wang2026cross}               & 77.8 & 94.0 & {53.7} & 73.1 & 74.9 & 91.4 & 92.2  & 93.8 \\
\midrule

\ourmethod                                   & \textbf{87.1} & \textbf{96.8} & \textbf{57.2} & \textbf{74.5} & \textbf{81.2} & \textbf{97.8} & 91.9  & {95.9} \\
\bottomrule
\end{tabular}%
}
\vspace{-.2cm}
\end{table}

\vspace{-.2cm}
\subsection{Ablation Study (RQ3)}
\vspace{-.1cm}

\begin{figure*}[t]
\centering
\vspace{-.3cm}
\begin{minipage}[t]{0.48\textwidth}
    \vspace{0pt}
    \centering

    \begin{minipage}[b][4.6cm][b]{\linewidth}
        \centering
        \includegraphics[width=\linewidth]
        {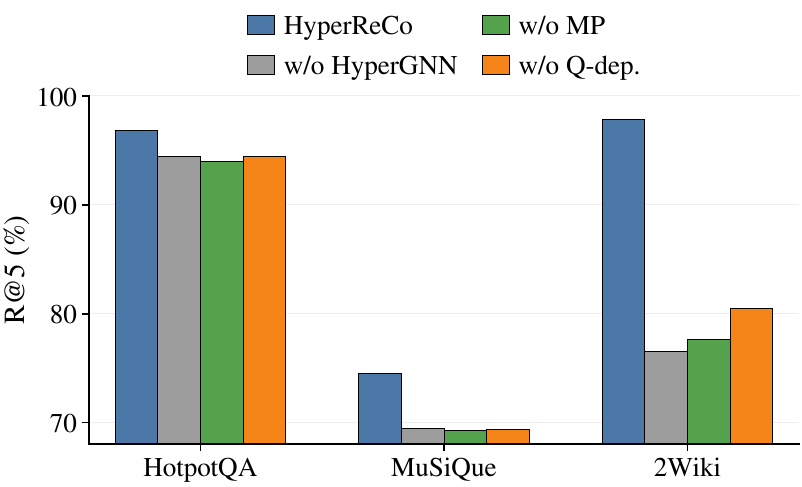}
    \end{minipage}

    \vspace{-0.15cm}
    \captionof{figure}{
        Ablation of structural relevance learning.
    }
    \label{fig:retrieval_ablation}
\end{minipage}
\hfill
\begin{minipage}[t]{0.48\textwidth}
    \vspace{0pt}
    \centering

    \begin{minipage}[b][4.6cm][b]{\linewidth}
        \centering
        \includegraphics[width=\linewidth]
        {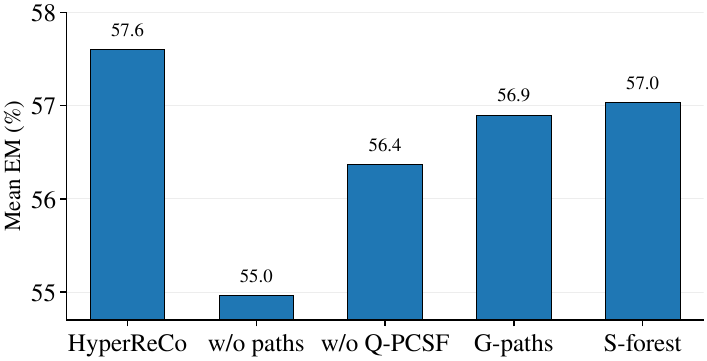}
    \end{minipage}

    \vspace{-0.15cm}
    \captionof{figure}{
        Analysis of hyper-path decoding strategies.
    }
    \label{fig:hyperpath_decoding}
\end{minipage}

\vspace{-0.5cm}
\end{figure*}

We conduct ablation studies to examine the contribution of query-conditioned structural relevance learning in the HyperGNN retriever. Specifically, we compare the full model with: (1) \emph{w/o HyperGNN}, which directly uses NV-Embed-v2 for retrieval; (2) \emph{w/o MP} (w/o message passing), which removes structural propagation between hyperedges and entities; and (3) \emph{w/o Q-dep} (w/o query-dependent modeling), which removes query conditioning from the HyperGNN. As shown in \Cref{fig:retrieval_ablation}, removing these components consistently degrades R@5 across the three multi-hop QA datasets, demonstrating that the retrieval gains arise from learning query-conditioned structural interactions rather than semantic relevance alone.


\begin{table*}[t]
\vspace{-.0cm}
\centering
\scriptsize
\setlength{\tabcolsep}{4pt}
\renewcommand{\arraystretch}{1.15}

\caption{Qualitative example of hyper-path-guided reasoning} 
\label{tab:app:case1}
\vspace{-0.3cm}
\begin{tabularx}{0.90\textwidth}{@{}lX@{}}
\toprule

\textbf{Question}
& When did Lord George Scott's father die? \\

\textbf{Gold Answer}
& 5 November 1914 \\

\textbf{Supporting Docs.}
& [``Lord George Scott'', ``William Montagu Douglas Scott, 6th Duke of Buccleuch''] \\

\textbf{Retrieved Docs.}
& [D1: \textbf{``Lord George Scott''},
D2: ``George Lyon, 5th Lord Glamis'',
D3: \textbf{``William Montagu Douglas Scott, 6th Duke of Buccleuch''},
D4: ``Charlotte Montagu Douglas Scott, Duchess of Buccleuch'',
D5: ``Walter Montagu Douglas Scott, 8th Duke of Buccleuch''] \\

\midrule

\textbf{Evidence Input}
& D1, D2, D3, D4, D5 \hfill \textbf{w/o Hyper-paths}\\

\textbf{Answer}
& The passage does not provide the date of Lord George Scott's father's death.
\hfill
\textcolor{red!80!black}{\ding{55} EM $=0$, F1 $=0$} \\

\midrule

\textbf{Evidence Input}
& D1, D2, D3, D4, D5 \hfill \textbf{\ourmethod}\\

\textbf{Decoded Hyper-paths}
&
Path 1: \textbf{D1 $\rightarrow$ [William Montagu Douglas Scott; 6th Duke of Buccleuch] $\rightarrow$ D3};
\\

\textbf{Answer}
& \textbf{5 November 1914}
\hfill
\textcolor{teal}{\ding{51} EM $=1$, F1 $=1$} \\

\bottomrule
\end{tabularx}
\vspace{-.5cm}
\end{table*}








\vspace{-.2cm}
\subsection{Analysis of Hyper-Path Effectiveness (RQ4)}
\label{sec:exp:case}
\vspace{-.1cm}

\noindent\textbf{Decoding Strategies.}
To examine whether the benefit of GGHD arises from identifying query-relevant evidence connections rather than merely providing structural paths, we compare the full GGHD with four variants: \emph{w/o paths}, which provides only the retrieved documents; \emph{w/o Q-PCSF}, which removes global forest optimization; \emph{G-paths}, a gradient-based path construction strategy; and \emph{S-forest}, a semantic-similarity-based forest construction strategy. As shown in \Cref{fig:hyperpath_decoding}, GGHD achieves the highest mean EM, demonstrating the benefit of combining gradient-guided connection scoring with global forest optimization. Detailed definitions of these variants are provided in Appendix~\ref{app:decoding-strategies} and we further provide an LLM-as-a-judge analysis in Appendix~\ref{app:hyperpath_judge} to assess whether GGHD produces more compact and useful hyper-paths.

\noindent\textbf{Case Study.}
As shown in~\Cref{tab:app:case1}, we further analyze a case where both settings retrieve the same top-5 documents with perfect supporting-document recall, but only \ourmethod produces the correct answer. Without hyper-path guidance, the LLM fails to associate the document identifying Lord George Scott's father with the document containing his death date. In contrast, GGHD explicitly exposes the key connection D1 $\rightarrow$ [William Montagu Douglas Scott, 6th Duke of Buccleuch] $\rightarrow$ D3, allowing the LLM to compose the two pieces of evidence and recover the correct answer. This example illustrates the central role of hyper-paths: beyond retrieving relevant documents, they explicitly identify how evidence distributed across documents should be connected for downstream reasoning. Full case details are provided in Appendix~\ref{sec:app:case}.

\subsection{Efficiency Analysis (RQ5)}
\label{sec:exp:efficiency}


\begin{wraptable}{r}{0.50\columnwidth}
\centering
\vspace{-0.5cm}
\caption{Efficiency comparison. Token cost is measured per 10K documents.
Time and EM are macro-averaged over HotpotQA, MuSiQue, and 2Wiki.}

\label{tab:efficiency}

\resizebox{0.48\columnwidth}{!}{%
\begin{tabular}{@{}lccc@{}}
\toprule
Method & Tokens / 10K $\downarrow$ & Time (s) $\downarrow$ & EM $\uparrow$ \\
\midrule
GraphRAG  & 76M & 5.02 & 37.7 \\
LightRAG  & 55M & 8.39 & 4.8 \\
GFM-RAG   & 48M & 3.36 & 52.1 \\
HGRAG     & \textbf{3.3M}             & \textbf{3.08} & 53.4 \\
\midrule
\ourmethod & \textbf{3.3M} & 3.55 & \textbf{57.6} \\
\bottomrule
\end{tabular}}

\vspace{-0.3cm}
\end{wraptable}

We evaluate the efficiency of \ourmethod in terms of offline index construction and online inference. As shown in~\Cref{tab:efficiency}, \ourmethod requires only 3.3M LLM tokens per 10K documents for index construction, substantially fewer than existing graph-based methods, as it avoids explicit relation extraction and summarization. For online inference, we report the macro-average end-to-end latency over HotpotQA, MuSiQue, and 2Wiki. \ourmethod takes 3.55 seconds per query, remaining comparable to GFM-RAG and HGRAG while achieving stronger QA performance. These results demonstrate that \ourmethod introduces limited inference overhead while enabling substantially lighter index construction. Detailed per-dataset latency and QA results are provided in Appendix~\ref{app:efficiency_detail}.

\section{Conclusion} \label{sec:conclusion}

We propose \ourmethod, a hypergraph-based framework for retrieving and connecting evidence for LLM multi-hop reasoning. A query-conditioned HyperGNN learns structural relevance over higher-order evidence associations to retrieve complementary evidence, while GGHD translates the learned interactions into explicit hyper-paths that guide cross-document evidence composition. Experiments on six benchmarks demonstrate strong retrieval and QA performance, with ablations and qualitative analyses confirming the effectiveness of both structural relevance learning and hyper-path guidance. Overall, our results show that successful multi-hop reasoning requires not only retrieving relevant evidence, but also exposing how that evidence is connected.


\section*{Declaration of Large Language Model Usage}
Generative AI tools were used to assist with proofreading and language polishing, as well as code implementation and debugging. LLMs were also used as experimental components for entity extraction, answer generation, and hyper-path evaluation are described in the corresponding method and experimental sections. All AI-assisted text was carefully reviewed and revised by the authors, and all AI-assisted code was inspected, tested, and verified for correctness before use. The authors take full responsibility for the accuracy, integrity, and final content of this work.

\section*{Ethics statement}
Our research addresses scientific questions only and does not involve human subjects, animals, or environmentally sensitive materials. Therefore, we do not anticipate any significant ethical risks or conflicts of interest. We are committed to maintaining high standards of scientific integrity and research ethics to ensure the validity, reliability, and transparency of our findings.

\section*{Reproducibility statement}
We provide detailed descriptions of the proposed method and its training and inference procedures in Section~\ref{sec:approach}. Experimental settings, implementation details, hyperparameters, and data processing procedures are provided in Section~\ref{sec:exp:exp-set} and Appendix~\ref{sec:app:methods}. Experimental settings and baselines have been rigorously verified to ensure fair comparison. We also specify the datasets, backbone models, retrieval settings, and language models used throughout the evaluation. Code and configurations for reproducing the main experiments will be made available.

\bibliography{iclr2027_conference}
\bibliographystyle{iclr2027_conference}

\appendix
\clearpage
\section{Datasets} \label{sec:appendix}

\subsection{Dataset Statistics}
\label{app:dataset_statistics}

\noindent\textbf{Training Datasets.}
We train \ourmethod{} on 6,000 training questions collected from HotpotQA, MuSiQue, and 2Wiki, following the training corpus used in GFM-RAG~\citep{luo2025gfm}. The resulting training corpus contains 53,573 documents and 326,286 extracted entities. Detailed statistics are shown in Table~\ref{tab:train_datasets}.

\begin{table}[h]
    \centering
    \small
    \setlength{\tabcolsep}{8pt}
    \caption{Statistics of the training datasets.}
    \label{tab:train_datasets}
    \begin{tabular}{lccc}
        \toprule
        Dataset & \#Questions & \#Documents & \#Entities \\
        \midrule
        HotpotQA & 2,000 & 19,527 & 130,484 \\
        MuSiQue  & 2,000 & 22,417 & 126,603 \\
        2Wiki    & 2,000 & 11,629 & 69,199 \\
        \midrule
        \textbf{Total}
        & \textbf{6,000}
        & \textbf{53,573}
        & \textbf{326,286} \\
        \bottomrule
    \end{tabular}
\end{table}

\noindent\textbf{Evaluation Datasets.}
We evaluate \ourmethod on three multi-hop QA benchmarks, including HotpotQA, MuSiQue, and 2Wiki, and three domain-specific GraphRAG benchmarks, including G-bench (Novel), G-bench (Medical), and G-bench (CS). The detailed statistics are summarized in Table~\ref{tab:test_datasets}.


\begin{table}[h]
    \centering
    \small
    \setlength{\tabcolsep}{8pt}
    \caption{Statistics of the evaluation datasets.}
    \label{tab:test_datasets}
    \begin{tabular}{@{}lcc@{}}
        \toprule
        Dataset                                       & \# Query & \# Document \\ \midrule
        HotpotQA \citep{yang2018hotpotqa}             & 1,000    & 9,221       \\
        MuSiQue \citep{trivedi2022musique}            & 1,000    & 11,656       \\
        2Wiki   \citep{ho2020constructing}            & 1,000    & 6,119      \\
        G-bench (Novel) \citep{xiang2025use}   & 2,010    & 461         \\
        G-bench (Medical) \citep{xiang2025use} & 2,062    & 2,406       \\
        G-bench (CS)   \citep{xiao2025graphrag} & 1,018      & 24,534      \\ \bottomrule
    \end{tabular}
\end{table}

\section{Methods} \label{sec:app:methods}

\subsection{HyperGraph Neural Network Retriever}
\subsubsection{Query-dependent Representation Initialization} \label{sec:app:hypergnn}
A frozen text encoder maps the query $q$, source document $d$, and entity $v$ into embeddings $x_q,x_d,x_v\in\mathbb{R}^{D}$, respectively. We project the query embedding into the HyperGNN hidden space as
\begin{equation}
    h_q=W_qx_q,
\end{equation}
where $W_q\in\mathbb{R}^{d_h\times D}$ is a learnable projection matrix.

For each document-induced hyperedge $e_d$, we first compute the query-document similarity $c_{qd}=\cos(x_q,x_d)$ and select a dense seed set $\mathcal{A}_q$ containing the top-$K$ documents according to $c_{qd}$. We define the normalized seed relevance as
\begin{equation}
    g_d=\begin{cases}
    \displaystyle \frac{\max(c_{qd},0)}{\max_{d'\in\mathcal{A}_q}\max(c_{qd'},0)+\epsilon}, & d\in\mathcal{A}_q,\\
    0, & \text{otherwise},
\end{cases}
\end{equation}
where $\epsilon$ is a small constant for numerical stability. The initial hyperedge state is then
\begin{equation}
    h_{e_d}^{(0)}=W_E x_d+\alpha_{\textrm{init}} g_d h_q,
\end{equation}
where $W_E$ projects the source-document embedding and $\alpha_{\textrm{init}}$ is a learnable scalar controlling the strength of query injection. Therefore, semantically relevant seed hyperedges receive stronger query conditioning before structural propagation.

For entities, we additionally incorporate query-linking and corpus-level structural features. Let $f_{v,q}\in\mathbb{R}^{F}$ denote the feature vector associated with entity $v$ under query $q$, including query-linking and entity-frequency information. We initialize the entity state as
\begin{equation}
    h_v^{(0)}=W_f f_{v,q}+\sigma(c_{qv})W_Vx_v,
\end{equation}
where $c_{qv}=\cos(x_q,x_v)$ and $\sigma(\cdot)$ denotes the sigmoid function. The query therefore conditions entity initialization through both $f_{v,q}$ and the query-entity similarity gate.

\subsubsection{Query-Conditioned Hypergraph Message Passing}
\label{sec:app:hypergnn-mp}

We provide the detailed message, aggregation, and update operators used by the HyperGNN. Let $q$ denote the input query, $d$ a source document, $e_d$ the hyperedge induced by $d$, and $v$ an entity contained in $e_d$. At propagation layer $\ell$, $h_{e_d}^{(\ell)}\in\mathbb{R}^{d_h}$ and $h_v^{(\ell)}\in\mathbb{R}^{d_h}$ denote the hidden states of hyperedge $e_d$ and entity $v$, respectively.

\paragraph{Query-Conditioned Roles.}
For each entity-hyperedge membership $v\in e_d$, we construct a query-conditioned role representation that characterizes how entity $v$ participates in hyperedge $e_d$ under query $q$. Let $x_q,x_d,x_v\in\mathbb{R}^{D}$ denote the frozen text-encoder representations of the query, source document, and entity name, respectively. We define the query-entity and query-document similarities as
\begin{equation}
    c_{qv}=\cos(x_q,x_v),\qquad c_{qd}=\cos(x_q,x_d),
\label{eq:app:mp-cosine}
\end{equation}
where $\cos(\cdot,\cdot)$ denotes cosine similarity.

We concatenate the semantic representations, membership features, and query similarities to form
\begin{equation}
    u_{v,e_d\mid q}=[x_v;x_d;x_q;\xi_{v,e_d};c_{qv};c_{qd}],
\label{eq:app:mp-role-input}
\end{equation}
where $[\cdot;\cdot]$ denotes vector concatenation and $\xi_{v,e_d}\in\mathbb{R}^{6}$ contains the auxiliary attributes associated with the membership $v\in e_d$, including the query-link indicator used in our implementation. Thus, $u_{v,e_d\mid q}\in\mathbb{R}^{3D+8}$ combines entity semantics, document semantics, query semantics, and membership-level information.

The role representation is then computed by
\begin{equation}
    z_{v,e_d\mid q}=\operatorname{LN}\left(W_{r,2}\operatorname{Dropout}\left(\operatorname{GELU}(W_{r,1}u_{v,e_d\mid q}+b_{r,1})\right)+b_{r,2}\right),
\label{eq:app:mp-role}
\end{equation}
where $W_{r,1}\in\mathbb{R}^{d_h\times(3D+8)}$ and $b_{r,1}\in\mathbb{R}^{d_h}$ project the concatenated input into the HyperGNN hidden space, while $W_{r,2}\in\mathbb{R}^{d_z\times d_h}$ and $b_{r,2}\in\mathbb{R}^{d_z}$ produce the final role representation $z_{v,e_d\mid q}\in\mathbb{R}^{d_z}$. $\operatorname{GELU}(\cdot)$ denotes the GELU activation, and $\operatorname{LN}(\cdot)$ denotes layer normalization. All projection matrices and bias terms are learned jointly with the HyperGNN.

\paragraph{Message Passing.}
At each layer $\ell$, the HyperGNN performs hyperedge-to-entity propagation followed by entity-to-hyperedge propagation. For each membership $v\in e_d$, we first construct direction-specific interaction representations:
\begin{equation}
u_{e_d\rightarrow v}^{(\ell)}=[h_{e_d}^{(\ell)};h_v^{(\ell)};z_{v,e_d\mid q}],\qquad u_{v\rightarrow e_d}^{(\ell)}=[h_v^{(\ell+1)};h_{e_d}^{(\ell)};z_{v,e_d\mid q}].
\label{eq:app:mp-inputs}
\end{equation}
Here, $u_{e_d\rightarrow v}^{(\ell)}$ represents the interaction used to propagate information from hyperedge $e_d$ to entity $v$, while $u_{v\rightarrow e_d}^{(\ell)}$ uses the updated entity state for the reverse propagation. Both representations concatenate the sender state, receiver state, and query-conditioned entity--hyperedge role $z_{v,e_d\mid q}$.

For each direction $r\in\{EV,VE\}$, where $EV$ and $VE$ denote hyperedge-to-entity and entity-to-hyperedge propagation, respectively, a two-layer MLP produces the message:
\begin{equation}
    \Psi_r^{(\ell)}(u)=W_{r,2}^{(\ell)}\operatorname{GELU}\left(W_{r,1}^{(\ell)}u+b_{r,1}^{(\ell)}\right)+b_{r,2}^{(\ell)}.
\label{eq:app:mp-message}
\end{equation}
A separate linear scoring function
\begin{equation}
    \psi_r^{(\ell)}(u)=(a_r^{(\ell)})^\top u+b_r^{(\ell)}
\label{eq:app:mp-attention-logit}
\end{equation}
produces a scalar attention logit for each directed message. Message and attention parameters are learned independently across propagation directions and layers.

For hyperedge-to-entity propagation, attention is normalized over all hyperedges containing the receiving entity $v$:
\begin{equation}
    \beta_{e_d\rightarrow v}^{(\ell)}=\frac{\exp\left(\psi_{EV}^{(\ell)}(u_{e_d\rightarrow v}^{(\ell)})\right)}{\sum_{d':\,v\in e_{d'}}\exp\left(\psi_{EV}^{(\ell)}(u_{e_{d'}\rightarrow v}^{(\ell)})\right)}.
\label{eq:app:mp-ev-attention}
\end{equation}
The resulting entity message is the attention-weighted aggregation
\begin{equation}
    m_v^{(\ell)}=\sum_{d:\,v\in e_d}\beta_{e_d\rightarrow v}^{(\ell)}\Psi_{EV}^{(\ell)}(u_{e_d\rightarrow v}^{(\ell)}).
\label{eq:app:mp-ev-agg}
\end{equation}
Therefore, an entity selectively aggregates information from the source hyperedges in which it occurs.

After updating the entity states, information is propagated back to the hyperedges. For entity-to-hyperedge propagation, attention is normalized over all entities contained in the receiving hyperedge $e_d$:
\begin{equation}
    \alpha_{v\rightarrow e_d}^{(\ell)}=\frac{\exp\left(\psi_{VE}^{(\ell)}(u_{v\rightarrow e_d}^{(\ell)})\right)}{\sum_{v'\in e_d}\exp\left(\psi_{VE}^{(\ell)}(u_{v'\rightarrow e_d}^{(\ell)})\right)}.
\label{eq:app:mp-ve-attention}
\end{equation}
The aggregated hyperedge message is
\begin{equation}
    m_{e_d}^{(\ell)}=\sum_{v\in e_d}\alpha_{v\rightarrow e_d}^{(\ell)}\Psi_{VE}^{(\ell)}(u_{v\rightarrow e_d}^{(\ell)}).
\label{eq:app:mp-ve-agg}
\end{equation}
This second propagation step allows each hyperedge to integrate query-conditioned information from its constituent entities, including information transferred through entities shared with other hyperedges.

\paragraph{State Updates.}
For both propagation directions, we update the receiving state using a GRU followed by residual addition, dropout, and layer normalization:
\begin{equation}
    \Phi_T^{(\ell)}(h,m)=\operatorname{LN}_T^{(\ell)}\left(h+\operatorname{Dropout}\left(\operatorname{GRU}_T^{(\ell)}(m,h)\right)\right),\qquad T\in\{V,E\}.
\label{eq:app:mp-update}
\end{equation}
Here, $T=V$ and $T=E$ denote entity and hyperedge updates, respectively, $m$ is the aggregated incoming message and $h$ is the previous receiver state. Within each layer, we first compute $h_v^{(\ell+1)}=\Phi_V^{(\ell)}(h_v^{(\ell)},m_v^{(\ell)})$ for all entities, and then use these updated entity states to compute $h_{e_d}^{(\ell+1)}=\Phi_E^{(\ell)}(h_{e_d}^{(\ell)},m_{e_d}^{(\ell)})$ for all hyperedges.

\subsection{Exact Q-PCSF Solver}
\label{sec:app:qpcsf}

We provide the exact optimization procedure used to solve the Q-PCSF objective in~\Cref{eq:approach:qpcsf}. Decoding is performed over the $K=5$ retrieved documents. For each candidate subset $F\subseteq E_q$, we first determine whether $(\mathcal{D}_q,F)$ is acyclic using a disjoint-set union structure.
Subsets containing cycles are discarded. For each valid forest, we evaluate
\begin{equation}
    \mathcal{J}(F)
    =
    \sum_{\{d_i,d_j\}\in F} c_{ij}
    +
    \sum_{d_j\in\mathcal{D}_q\setminus\{d_a\}}
    \pi_{aj}
    \mathbb{1}[d_a\not\leftrightarrow_F d_j], \label{eq:app:qpcsf}
\end{equation}

The \Cref{eq:app:qpcsf} is identical to the objective in~\Cref{eq:approach:qpcsf}. Connectivity is evaluated over the entire selected forest, so a demand is considered satisfied whenever $d_a$ and $d_j$ are connected either directly or through intermediate retrieved documents. The empty forest is also included as a valid candidate, with objective
\begin{equation}
    \mathcal{J}(\varnothing)
    =
    \sum_{d_j\in\mathcal{D}_q\setminus\{d_a\}}
    \pi_{aj}.
\end{equation}
The exact solver proceeds as follows:
\begin{enumerate}
    \item Initialize the best solution with the empty forest.
    \item Enumerate all nonempty subsets $F\subseteq E_q$ and discard those containing cycles.
    \item For each valid forest, compute $\mathcal{J}(F)$ using the selected-edge costs and the penalties of unsatisfied connectivity demands.
    \item Return the forest with the minimum objective value. Exact ties are resolved by preferring fewer selected edges and then the lexicographically smaller sequence of document-index pairs.
\end{enumerate}

\subsection{Training Objectives}
\label{sec:app:optimization}

Given a query $q$, let $\mathcal{S}_q\subseteq\mathcal{D}$ denote its supporting documents. We assign each document-induced hyperedge $e_d$ a binary relevance label $y_d=\mathbb{1}[d\in\mathcal{S}_q]$. Following GFM-RAG~\citep{luo2025gfm}, we further derive weak entity supervision from the supporting documents:
\begin{equation}
    \mathcal{V}_q^{+}=\bigcup_{d\in\mathcal{S}_q}e_d,\qquad y_v=\mathbb{1}[v\in\mathcal{V}_q^{+}].
\end{equation}
Let $p_d=\sigma(s_d)$ and $p_v=\sigma(s_v)$ denote the predicted relevance scores of documents and entities, respectively.

\paragraph{Document Relevance.}
We supervise document relevance using binary cross-entropy:
\begin{equation}
    \mathcal{L}_{\mathrm{ret}}^{d}=-\frac{1}{|\mathcal{D}|}\sum_{d\in\mathcal{D}}\left[y_d\log p_d+(1-y_d)\log(1-p_d)\right].
\label{eq:app:doc-ret}
\end{equation}
Following GFM-RAG~\citep{luo2025gfm}, we additionally employ a ranking loss $\mathcal{L}_{\mathrm{rank}}^{d}$ to encourage supporting documents to receive higher relevance scores than negative documents.

\paragraph{Entity Relevance.}
For entity supervision, we combine pointwise and listwise relevance learning:
\begin{equation}
    \mathcal{L}_{\mathrm{ret}}^{e}=\alpha\mathcal{L}_{\mathrm{BCE}}^{e}+(1-\alpha)\mathcal{L}_{\mathrm{list}}^{e},
\label{eq:app:entity-ret}
\end{equation}
where $\mathcal{L}_{\mathrm{BCE}}^{e}$ supervises binary entity relevance and $\mathcal{L}_{\mathrm{list}}^{e}$ encourages positive entities to receive higher relevance mass over the complete entity set. Following the GFM-RAG training setup~\citep{luo2025gfm}, we set $\alpha=0.3$. The complete retrieval objective is
\begin{equation}
\mathcal{L}_{\mathrm{ret}}=\mathcal{L}_{\mathrm{ret}}^{d}+\lambda_{\mathrm{rank}}\mathcal{L}_{\mathrm{rank}}^{d}+\lambda_{\mathrm{ent}}\mathcal{L}_{\mathrm{ret}}^{e}.
\label{eq:app:ret-loss}
\end{equation}

\paragraph{Semantic Distillation.}
To prevent structural learning from drifting away from the semantic relevance captured by the frozen text encoder, we further introduce a semantic distillation objective. We use the frozen text encoder as a semantic teacher to regularize document relevance prediction. Specifically, the encoder similarity and HyperGNN logit are converted into Bernoulli relevance scores,
\begin{equation}
    p_{\mathrm{enc}}(d\mid q)=\sigma(c_{qd}),\qquad p_{\mathrm{hyp}}(d\mid q)=\sigma(s_d),
\end{equation}
and we minimize their KL divergence:
\begin{equation}
    \mathcal{L}_{\mathrm{KL}}=\sum_{d\in\mathcal{D}}D_{\mathrm{KL}}\left(\operatorname{Bern}(p_{\mathrm{enc}}(d\mid q))\,\Vert\,\operatorname{Bern}(p_{\mathrm{hyp}}(d\mid q))\right).
\label{eq:app:kl-loss}
\end{equation}

The final optimization objective is
\begin{equation}
\theta^\star=\operatorname*{arg\,min}_{\theta}\mathbb{E}_{q\sim\mathcal{Q}_{\mathrm{train}}}\left[\mathcal{L}_{\mathrm{ret}}(\theta;q)+\lambda_{\mathrm{KL}}\mathcal{L}_{\mathrm{KL}}(\theta;q)\right].
\label{eq:app:overall-objective}
\end{equation}

\subsection{Implementation Details}
\label{sec:app:implementation}

\noindent\textbf{Hypergraph Construction.}
Following HGRAG~\citep{wang2026cross}, we use the same LLM-based entity extraction prompt to identify entities from each document. The extracted mentions are normalized into canonical entities and used to construct the hypergraph as described in Preliminary. Since we directly follow the entity extraction protocol of HGRAG, we do not introduce additional prompting strategies for hypergraph construction.

\noindent\textbf{Model Settings.}
We use frozen NV-Embed-v2 representations with an embedding dimension of 4096 for questions, documents, and entities. The HyperGNN uses a hidden dimension of 384, an entity-hyperedge role dimension of 192, and two message-passing layers. We use five dense seed documents and apply dropout with rate 0.1. Both the document and entity scoring heads use two-layer MLPs with hidden dimension 384 and GELU activation.

\noindent\textbf{Training Settings.}
We train on 6,000 questions from HotpotQA, MuSiQue, and 2Wiki. The model is optimized with AdamW using a learning rate of $3\times10^{-4}$, weight decay of $10^{-2}$, and $(\beta_1,\beta_2)=(0.9,0.999)$. Training is performed on two NVIDIA RTX 4090 GPUs with an effective batch size of 2. We train for at most 30 epochs with early-stopping patience 5 and random seed 42. The ranking margin is set to 1.0 with at most 128 hard negatives. Following the official training configuration of GFM-RAG~\citep{luo2025gfm}, we set the weights of the entity BCE and listwise objectives to 0.3 and 0.7, respectively, with an adversarial temperature of 0.2 for negative reweighting. Detailed training objectives are provided in Appendix~\ref{sec:app:optimization}.

\subsection{Detailed Prompts for Inference}
\label{sec:app:prompt}
\begin{figure}[htb]
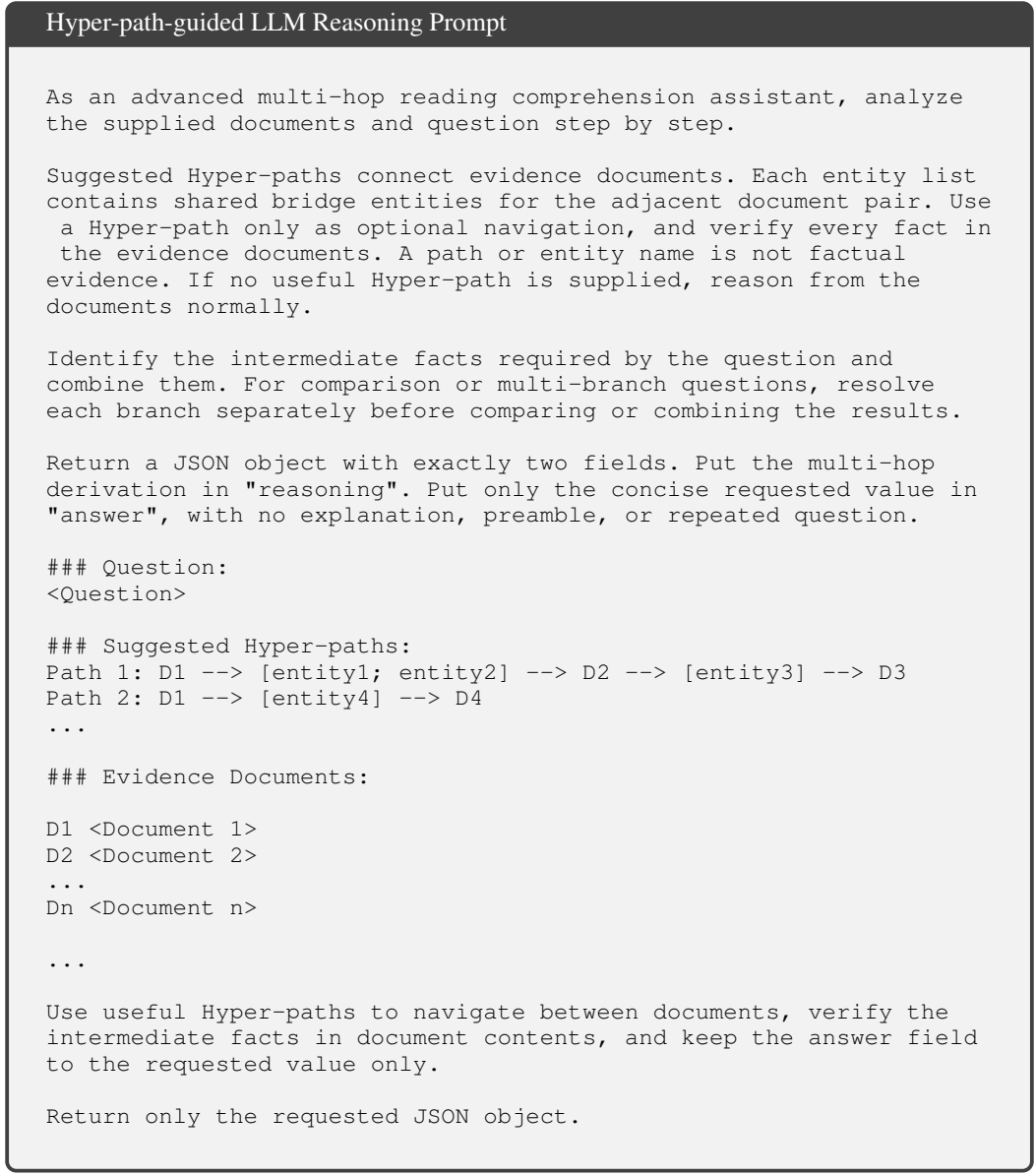

    \centering
    \begin{minipage}{0.99\columnwidth}
        \centering
        \begin{tcolorbox}[title=Hyper-path-guided LLM Reasoning Prompt]
            \small
            \begin{lstlisting}
As an advanced multi-hop reading comprehension assistant, analyze the supplied documents and question step by step.

Suggested Hyper-paths connect evidence documents. Each entity list contains shared bridge entities for the adjacent document pair. Use a Hyper-path only as optional navigation, and verify every fact in the evidence documents. A path or entity name is not factual evidence. If no useful Hyper-path is supplied, reason from the documents normally.

Identify the intermediate facts required by the question and combine them. For comparison or multi-branch questions, resolve each branch separately before comparing or combining the results.

Return a JSON object with exactly two fields. Put the multi-hop derivation in "reasoning". Put only the concise requested value in "answer", with no explanation, preamble, or repeated question.

### Question:
<Question>

### Suggested Hyper-paths:
Path 1: D1 --> [entity1; entity2] --> D2 --> [entity3] --> D3
Path 2: D1 --> [entity4] --> D4
...

### Evidence Documents:

D1 <Document 1>
D2 <Document 2>
...
Dn <Document n>

...

Use useful Hyper-paths to navigate between documents, verify the intermediate facts in document contents, and keep the answer field to the requested value only.

Return only the requested JSON object.
            \end{lstlisting}
        \end{tcolorbox}
        \vspace{1mm}
    \end{minipage}
    \caption{The prompt template used in \ourmethod for hyper-path-guided LLM reasoning.}
    \label{fig:reasoning_prompt_hyperpath}
\end{figure}

For QA inference, we follow the prompting protocols adopted by prior GraphRAG systems such as GFM-RAG~\citep{luo2025gfm} and HGRAG~\citep{wang2026cross} for the baseline methods. For \ourmethod, we additionally provide the decoded hyper-paths alongside the retrieved documents as explicit cross-document navigation cues, while the document contents remain the factual evidence for answer generation. The complete inference prompt used by \ourmethod is shown in Figure~\ref{fig:reasoning_prompt_hyperpath}.

\subsection{Detailed Descriptions of Baseline Methods}
\label{sec:app:baseline}

We compare \ourmethod with two groups of representative retrieval methods: \textit{non-structure methods} and \textit{graph-enhanced methods}. The detailed descriptions of the baselines are as follows.

\paragraph{Non-structure Methods.}
These methods retrieve documents mainly based on lexical or semantic relevance without explicitly modeling cross-document structures.

\begin{itemize}
    \item \textbf{BM25}~\citep{robertson1994some} is a classical sparse retrieval method that ranks documents according to query-term statistics based on the probabilistic retrieval model.

    \item \textbf{ColBERTv2}~\citep{santhanam2022colbertv2} is a dense retriever based on token-level late interaction, together with residual compression and denoised supervision for effective and efficient retrieval.

    \item \textbf{Qwen3-Embedding-8B}~\citep{zhang2025qwen3} is an LLM-based embedding model built on Qwen3 and trained with a multi-stage pipeline for general-purpose semantic retrieval.

    \item \textbf{NV-Embed-v2}~\citep{lee2025nv} is a general-purpose LLM-based embedding model that employs latent-attention pooling and contrastive instruction tuning to produce strong representations for dense retrieval.
\end{itemize}

\paragraph{Graph-enhanced Methods.}
These methods introduce graph structures to capture dependencies beyond independent document similarity.

\begin{itemize}
    \item \textbf{RAPTOR}~\citep{sarthiraptor} recursively embeds, clusters, and summarizes text chunks to construct a hierarchical tree, enabling retrieval at different levels of abstraction.

    \item \textbf{GraphRAG}~\citep{edge2024local} constructs an entity-based knowledge graph and applies hierarchical community detection to generate community summaries for retrieval and LLM generation.

    \item \textbf{LightRAG}~\citep{guo2024lightrag} incorporates graph structures into text indexing and employs dual-level retrieval to access both low-level entities and high-level semantic information.

    \item \textbf{HippoRAG}~\citep{jimenez2024hipporag} constructs a knowledge graph from the corpus and applies Personalized PageRank to propagate query relevance for associative multi-hop retrieval.

    \item \textbf{HippoRAG~2}~\citep{gutierrez2025ragmemory} extends HippoRAG by more deeply integrating passage nodes and semantic passage signals into graph-based retrieval.

    \item \textbf{GFM-RAG}~\citep{luo2025gfm} trains a query-dependent GNN through graph completion pre-training and supervised retrieval fine-tuning to capture query--knowledge relationships over knowledge graphs.

    \item \textbf{G-retriever}~\citep{he2024g} formulates graph retrieval as a prize-collecting Steiner tree optimization problem to select query-relevant subgraphs for LLM-based question answering.

    \item \textbf{Quest-GNN}~\citep{yan2026questgnn} constructs a multi-information-level knowledge graph and performs question-guided intra- and inter-level message passing for multi-hop retrieval.

    \item \textbf{EHRAG}~\citep{song2026ehrag} constructs a hybrid hypergraph with structural and semantic hyperedges and retrieves evidence using structure--semantic hypergraph diffusion with relevance refinement.

    \item \textbf{HGRAG}~\citep{wang2026cross} represents entities as nodes and source passages as hyperedges, integrating entity- and passage-level semantic relevance through hypergraph diffusion for multi-hop retrieval.
\end{itemize}

\section{Additional Experiments and Analyses}

\begin{table*}[t]
\centering
\small
\setlength{\tabcolsep}{5pt}
\caption{Retrieval performance with different embedding models.
We report document Recall@2 (R@2) and Recall@5 (R@5).
``Embedding'' denotes direct dense retrieval using the corresponding
embedding model.}
\label{tab:embedding_ablation}
\begin{tabular}{llcccccc}
\toprule
\multirow{2}{*}{Embedding Model}
& \multirow{2}{*}{Method}
& \multicolumn{2}{c}{HotpotQA}
& \multicolumn{2}{c}{MuSiQue}
& \multicolumn{2}{c}{2Wiki} \\

\cmidrule(lr){3-4}
\cmidrule(lr){5-6}
\cmidrule(lr){7-8}

& & R@2 & R@5 & R@2 & R@5 & R@2 & R@5 \\

\midrule

\multirow{3}{*}{NV-Embed-v2 (7B) \citep{lee2025nv}}
& Embedding & 84.1 & 94.4 & 52.7 & 69.5 & 69.1 & 76.5 \\
& HGRAG     & 77.8 & 94.0 & 53.7 & 73.1 & 74.9 & 91.4 \\
& \textbf{\ourmethod}
              & \textbf{87.1} & \textbf{96.8}
              & \textbf{57.2} & \textbf{74.5}
              & \textbf{81.2} & \textbf{97.8} \\

\midrule

\multirow{3}{*}{Qwen3-Emb (8B) \citep{zhang2025qwen3}}
& Embedding & 74.1 & 88.8 & 46.8 & 62.1 & 66.2 & 74.1 \\
& HGRAG     & 72.2 & 88.0 & 47.2 & 61.1 & 66.3 & 74.1 \\
& \textbf{\ourmethod}
              & \textbf{82.3} & \textbf{95.0}
              & \textbf{52.6} & \textbf{69.3}
              & \textbf{80.7} & \textbf{97.6} \\

\midrule

\multirow{3}{*}{Qwen3-Emb (4B) \citep{zhang2025qwen3}}
& Embedding & 71.6 & 86.3 & 44.5 & 59.0 & 66.1 & 72.7 \\
& HGRAG     & 70.7 & 86.4 & 44.4 & 58.8 & 65.6 & 72.5 \\
& \textbf{\ourmethod}
              & \textbf{81.5} & \textbf{94.3}
              & \textbf{51.2} & \textbf{68.9}
              & \textbf{80.8} & \textbf{97.4} \\

\midrule

\multirow{3}{*}{Qwen3-Emb (0.6B) \citep{zhang2025qwen3}}
& Embedding & 66.5 & 79.2 & 41.7 & 54.3 & 64.6 & 70.3 \\
& HGRAG     & 65.7 & 79.3 & 41.3 & 54.0 & 64.0 & 69.8 \\
& \textbf{\ourmethod}
              & \textbf{78.5} & \textbf{91.9}
              & \textbf{50.3} & \textbf{66.5}
              & \textbf{79.6} & \textbf{97.0} \\

\bottomrule
\end{tabular}
\end{table*}
\subsection{Robustness to Different Embedding Models}
\label{app:embedding-robustness}
As shown in~\Cref{tab:embedding_ablation}, we further examine whether the performance of \ourmethod depends on a particular embedding model. We replace the default NV-Embed-v2 encoder with Qwen3-Embedding models of different scales (8B, 4B, and 0.6B), while keeping all other components and experimental settings unchanged. \ourmethod maintains strong retrieval performance across all embedding backbone models, even when the encoder size is substantially reduced. For example, on 2Wiki, R@5 remains at or above 97\% across all four embedding models, while on HotpotQA it remains above 91\% even with the 0.6B encoder. Moreover, \ourmethod consistently outperforms HGRAG under every embedding setting. These results demonstrate that \ourmethod is largely agnostic to the choice of query encoder as it remains effective across embedding models of different scales.

\subsection{Detailed Inference Efficiency}
\label{app:efficiency_detail}
\begin{table*}[t]
\centering
\small
\caption{
    Detailed inference efficiency across three multi-hop QA datasets. Avg. denotes the macro-average across HotpotQA, MuSiQue, and 2Wiki.
}
\label{tab:app_efficiency}

\resizebox{.95\textwidth}{!}{%
\begin{tabular}{@{}lcccccccc@{}}
\toprule
& \multicolumn{2}{c}{HotpotQA}
& \multicolumn{2}{c}{MuSiQue}
& \multicolumn{2}{c}{2Wiki}
& \multicolumn{2}{c}{Avg.} \\
\cmidrule(lr){2-3}
\cmidrule(lr){4-5}
\cmidrule(lr){6-7}
\cmidrule(lr){8-9}
Method
& Time (s) $\downarrow$ & EM $\uparrow$
& Time (s) $\downarrow$ & EM $\uparrow$
& Time (s) $\downarrow$ & EM $\uparrow$
& Time (s) $\downarrow$ & EM $\uparrow$ \\
\midrule
GraphRAG
& 4.10 & 51.4
& 6.18 & 27.0
& 4.79 & 34.7
& 5.02 & 37.7 \\

LightRAG
& 8.71 & 9.9
& 8.56 & 2.0
& 7.89 & 2.5
& 8.39 & 4.8 \\

GFM-RAG
& 3.83 & 56.2
& \textbf{3.27} & 30.2
& \textbf{2.99} & 69.8
& 3.36 & 52.1 \\

HGRAG
& \textbf{2.92} & 55.9
& 3.33 & 36.2
& 2.99 & 68.2
& \textbf{3.08} & 53.4 \\
\midrule
\ourmethod
& 3.34 & \textbf{60.7}
& 4.01 & \textbf{39.9}
& 3.30 & \textbf{72.2}
& 3.55 & \textbf{57.6} \\
\bottomrule
\end{tabular}}
\end{table*}

To provide a more detailed view of online efficiency, we report the end-to-end inference latency and corresponding QA performance on each multi-hop QA dataset in~\Cref{tab:app_efficiency}. The end-to-end time includes both retrieval and downstream QA inference.

As shown in~\Cref{tab:app_efficiency}, \ourmethod maintains consistent end-to-end efficiency across all three datasets, with an average latency of 3.55 seconds per query. Its latency is close to HGRAG (3.08 s) and GFM-RAG (3.36 s), while being substantially lower than GraphRAG (5.02 s) and LightRAG (8.39 s). More importantly, \ourmethod achieves the highest EM on all three datasets, improving the macro-average EM by 4.2 points over HGRAG and 5.5 points over GFM-RAG. These results indicate that the additional structural reasoning and hyper-path decoding introduce only moderate computational overhead while providing consistent gains in downstream QA performance.

\begin{table}[t]
\centering
\caption{
Detailed EM (\%) results for different hyper-path decoding strategies. Mean denotes the macro average over HotpotQA, MuSiQue, and 2Wiki.
}
\label{tab:decoding_strategy_full}

\setlength{\tabcolsep}{5.5pt}
\begin{tabular}{lcccc}
\toprule
Method
& HotpotQA
& MuSiQue
& 2Wiki
& Mean \\
\midrule

\textbf{HyperReCo}
& \textbf{60.7}
& \textbf{39.9}
& \textbf{72.2}
& \textbf{57.6} \\

\emph{w/o paths}
& 57.7
& 35.6
& 71.6
& 55.0 \\

\emph{w/o Q-PCSF}
& 59.8
& 38.5
& 70.8
& 56.4 \\

G-paths
& 59.9
& 38.8
& 72.0
& 56.9 \\

S-forest
& 60.2
& 39.4
& 71.5
& 57.0 \\

\bottomrule
\end{tabular}
\end{table}
\subsection{Additional Analysis of Hyper-Path Decoding}
\label{app:decoding-strategies}

We provide additional details of the hyper-path decoding variants evaluated in
\Cref{fig:hyperpath_decoding}. 
For all variants, we keep the retrieved document set, downstream reader, and answer-generation procedure unchanged, and modify only how the connections among retrieved documents are constructed and presented to the reader. This isolates the contribution of gradient-guided connection scoring and Q-PCSF-based global structure optimization.

\paragraph{HyperReCo (GGHD).}
Our full decoder derives query-dependent connection scores from retriever gradients, converts them into costs, and applies Q-PCSF to select a compact forest over the retrieved documents. The forest is then expanded into hyper-paths and provided to the reader.

\paragraph{\emph{w/o paths}.}
This variant removes hyper-path decoding and provides only the retrieved documents to the reader, measuring the benefit of explicitly exposing evidence connections.

\paragraph{\emph{w/o Q-PCSF}.}
This variant keeps the gradient-based connection scores but removes global Q-PCSF optimization, selecting connections only from local scores.

\paragraph{\emph{G-paths}.}
G-paths directly constructs evidence paths from gradient-based importance scores, following the path extraction principle of NBFNet-style reasoning, without global forest optimization.

\paragraph{\emph{S-forest}.}
S-forest keeps the forest-based decoding procedure but replaces gradient-guided connection costs with semantic-similarity-based costs.

As shown in \Cref{tab:decoding_strategy_full}, removing explicit hyper-paths leads to the largest degradation, decreasing the mean EM from 57.6 to 55.0, which confirms that making evidence connections explicit is beneficial to the reader. Restoring explicit evidence connections improves performance, while both G-paths and S-forest remain below the full GGHD. These results suggest that the gain is not explained simply by adding arbitrary structural paths: both query-dependent connection scoring and global forest optimization contribute to the final performance.

\subsection{LLM-as-a-Judge Evaluation of Hyper-Paths}
\label{app:hyperpath_judge}

\begin{figure}[htb]
    \centering

    \begin{minipage}{0.99\columnwidth}
        \centering
        \begin{tcolorbox}[title=Hyper-path Helpfulness Judge Prompt]
            \footnotesize
            \begin{lstlisting}
You are judging whether each supplied hyper-path helps answer a question.
Use only the question, reference answer, path order, bridge entities, and supplied document text. Do not use outside knowledge. The reference answer is the target claim to verify, not evidence. A shared entity, graph connection, or topical similarity alone does not make a path helpful.

Assess the paths as a group first so you can recognize complementary paths. Then label each path exactly once:
- helpful=1 only when its document text provides evidence for a claim in the reference answer, or a necessary intermediate fact/relation in a coherent route toward that answer. A path may be helpful even if it supplies only one necessary step and the full set still has a gap.
- helpful=0 when it is off-topic, connected only by a superficial shared entity, or does not contribute evidence to answering the question. Redundancy alone is not a reason for 0 if the path independently supports or corroborates a relevant claim.

            \end{lstlisting}
        \end{tcolorbox}
        \vspace{1mm}
    \end{minipage}
    \caption{The prompt template used for LLM-as-a-judge evaluation of hyper-path helpfulness.}
    \label{tab:judge_prompt}
\end{figure}

\paragraph{Evaluation protocol.}
We use GPT-4o as an LLM judge to assess whether decoded hyper-paths contribute text-grounded evidence toward answering a question. We randomly sample 100 questions from each of HotpotQA, MuSiQue, and 2Wiki, resulting in 300 questions in total for evaluation. For each question, the judge receives the question, reference answer, and the paths produced by a decoding strategy, including their document order, bridge entities, and associated document text. We apply the same evaluation prompt to GGHD, \emph{w/o Q-PCSF}, \emph{G-paths}, and \emph{S-forest}, with all variants described in detail in Section~\ref{app:decoding-strategies}. The detailed judging prompt is provided in~\Cref{tab:judge_prompt}.

\paragraph{Helpfulness criterion.}
The judge first considers the paths jointly to identify complementary evidence, then assigns each path a binary helpfulness label. A hyper-path is considered helpful if it establishes an evidence-supported connection between retrieved documents that facilitates multi-hop reasoning toward the reference answer.

\paragraph{Metrics.}
We report the \emph{helpful-path ratio} and the \emph{mean number of paths per question}, measuring judged utility and output compactness, respectively. Let $K_q$ and $U_q$ denote the numbers of supplied and helpful paths for question $q$. Over the evaluated question set $\mathcal{Q}$, the two metrics are
\begin{equation}
    \mathrm{HelpfulRatio}
    =
    \frac{100}{|\mathcal{Q}|}
    \sum_{q\in\mathcal{Q}}
    \frac{U_q}{\max(K_q,1)},
    \qquad
    \mathrm{MeanPaths}
    =
    \frac{1}{|\mathcal{Q}|}
    \sum_{q\in\mathcal{Q}} K_q.
\end{equation}

\begin{table}[t]
    \centering
    \caption{
        LLM-judged hyper-path utility and compactness.
        Helpful-path ratios are averaged over questions.
        GGHD achieves the highest helpful-path ratio with the fewest
        paths per question.
    }
    \label{tab:hyperpath_judge}
    \setlength{\tabcolsep}{8pt}
    \begin{tabular}{lcc}
        \toprule
        Method
        & \shortstack{Helpful-path ratio\\(\%) $\uparrow$}
        & \shortstack{Mean paths\\per question $\downarrow$} \\
        \midrule
        \textbf{GGHD}
        & \textbf{59.2}
        & \textbf{1.68} \\
        \emph{w/o Q-PCSF}
        & 28.4
        & 5.39 \\
        \emph{G-paths}
        & 43.1
        & 2.97 \\
        \emph{S-forest}
        & 50.6
        & 2.11 \\
        \bottomrule
    \end{tabular}
\end{table}

\paragraph{Results and analysis.}
As shown in \Cref{tab:hyperpath_judge}, GGHD achieves the highest helpful-path ratio (59.2\%) with the fewest paths per question (1.68). Removing Q-PCSF substantially increases the number of paths while reducing their helpfulness, suggesting that global forest optimization improves path selectivity. GGHD also outperforms G-paths and S-forest while generating fewer paths, indicating that it produces more compact and answer-relevant evidence connections.

\subsection{Qualitative Analysis of Decoded Hyper-Paths}
\label{sec:app:case}

\Cref{sec:app:case1_details} provides the complete decoded hyper-paths for the qualitative example discussed in Section~\ref{sec:exp:case}. The question asks for the death date of Lord George Scott's father. The required reasoning spans two supporting documents: D1 identifies \emph{William Montagu Douglas Scott, 6th Duke of Buccleuch} as Lord George Scott's father, while D3 provides his death date as \emph{5 November 1914}. Without hyper-path guidance, these two facts remain implicitly distributed across the retrieved documents, and the LLM fails to associate the entity identified in D1 with the information provided in D3.

Among the decoded routes, Path~1,
\[
\mathrm{D1}
\rightarrow
[\text{William Montagu Douglas Scott, 6th Duke of Buccleuch}]
\rightarrow
\mathrm{D3},
\]
makes the required cross-document dependency explicit. Importantly, the hyper-path does not introduce any additional factual information, and it only indicates that D1 and D3 are connected through a shared entity. This structural cue provides the LLM with a direct navigation route from identifying the father in D1 to locating the corresponding death date in D3, leading to the correct answer. The result supports our motivation that successful multi-hop reasoning depends not only on retrieving the necessary evidence, but also on making the connections required for evidence composition explicit.

\begin{table*}[!htbp]
    \centering
    \renewcommand{\arraystretch}{1.3}
    \caption{Qualitative case study of cross-document evidence composition with and without hyper-path guidance.
Both settings retrieve all supporting documents (Recall@5 = 1.0), yet only the hyper-path-guided setting produces the correct answer.}
    \label{sec:app:case1_details}
    \small
    \begin{tabularx}{0.95\textwidth}{@{}l|X@{}}
        \toprule
        \textbf{Question}
        & When did Lord George Scott's father die? \\

        \textbf{Gold Answer}
        & 5 November 1914 \\

        \textbf{Supporting Docs.}
        & [``Lord George Scott'', ``William Montagu Douglas Scott, 6th Duke of Buccleuch''] \\

        \textbf{Retrieved Docs.}
        & [D1: \textbf{``Lord George Scott''},
        D2: ``George Lyon, 5th Lord Glamis'',
        D3: \textbf{``William Montagu Douglas Scott, 6th Duke of Buccleuch''},
        D4: ``Charlotte Montagu Douglas Scott, Duchess of Buccleuch'',
        D5: ``Walter Montagu Douglas Scott, 8th Duke of Buccleuch'']
        \hfill
        \textcolor{teal}{\textbf{(Recall@5 = 1.0)}} \\
        \midrule

        \multicolumn{2}{c}{\textbf{w/o Hyper-paths}} \\
        \midrule

        \textbf{LLM Input}
        & \textbf{Question:} When did Lord George Scott's father die?
        \newline
        \textbf{Retrieved Docs:} D1, D2, D3, D4, D5 \\

        \textbf{Reasoning}
        & The passage provides information about Lord George Scott and his family,
        specifically mentioning that he was the son of William Montagu Douglas Scott,
        6th Duke of Buccleuch. However, it does not provide a specific date of death
        for Lord George Scott's father. The only date mentioned in the passage is the
        death date of Lord George Scott himself, which is 23 February 1947. Since the
        passage does not include any information about the death of William Montagu
        Douglas Scott, 6th Duke of Buccleuch, I cannot determine when he died based
        on the provided text. \\

        \textbf{Answer}
        & The passage does not provide the date of Lord George Scott's father's death.
        \hfill
        \textcolor{red!80!black}{
        \ding{55} \textbf{EM = 0.0, F1 = 0.0}} \\
        \midrule

        \multicolumn{2}{c}{\textbf{\ourmethod}} \\
        \midrule

        \textbf{LLM Input}
        & \textbf{Question:} When did Lord George Scott's father die?
        \newline
        \textbf{Suggested Hyper-paths:}\newline
        Path 1: \textbf{D1 $\rightarrow$ [William Montagu Douglas Scott; 6th Duke of Buccleuch] $\rightarrow$ D3};\newline
        Path 2: D1 $\rightarrow$ [John Montagu Douglas Scott; 7th Duke of Buccleuch] $\rightarrow$ D5;\newline
        \newline
        \textbf{Evidence Docs:} D1, D2, D3, D4, D5 \\

        \textbf{Reasoning}
        & To find the death date of Lord George Scott's father, we identify him
        as William Montagu Douglas Scott, 6th Duke of Buccleuch from D1.
        D3 provides the death date of William Montagu Douglas Scott as
        5 November 1914. Therefore, the answer is 5 November 1914. \\

        \textbf{Answer}
        & 5 November 1914
        \hfill
        \textcolor{teal}{
        \ding{51} \textbf{EM = 1.0, F1 = 1.0}} \\
        \bottomrule
    \end{tabularx}
\end{table*}

\end{document}